\documentclass[10pt,twocolumn,letterpaper]{article}

\usepackage[pagenumbers]{cvpr} 

\usepackage[dvipsnames]{xcolor}

\usepackage{color}
\usepackage{wrapfig}
\usepackage{multirow}

\definecolor{cvprblue}{rgb}{0.21,0.49,0.74}
\definecolor{mypink}{RGB}{239,43,159}
\usepackage[pagebackref,breaklinks,colorlinks,citecolor=cvprblue]{hyperref}
\usepackage{xcolor,colortbl}

\def\paperID{*****} 
\def\confName{CVPR}
\def\confYear{2024}

\title{NExT-Chat: An LMM for Chat, Detection and Segmentation}

\author{
Ao Zhang$^{1}$ \ \  
Yuan Yao$^{1*}$ \ \ \
Wei Ji$^1$ \ \
Zhiyuan Liu$^{2}$ \ \ 
Tat-Seng Chua$^1$
\\[0.5em]
$^1$National University of Singapore \ \ \
$^2$Tsinghua University \ \ \ 
\ \ \ \\
{\small \texttt{aozhang@u.nus.edu} \quad \texttt{yaoyuanthu@gmail.com}} \\
\\
\large{\color{mypink} \textbf{ \url{https://next-chatv.github.io}}}
}

\begin{document}
\twocolumn[{%
\renewcommand\twocolumn[1][]{#1}%

\maketitle

\begin{center}
    \centering
    \captionsetup{type=figure}
    \includegraphics[width=0.8\textwidth]{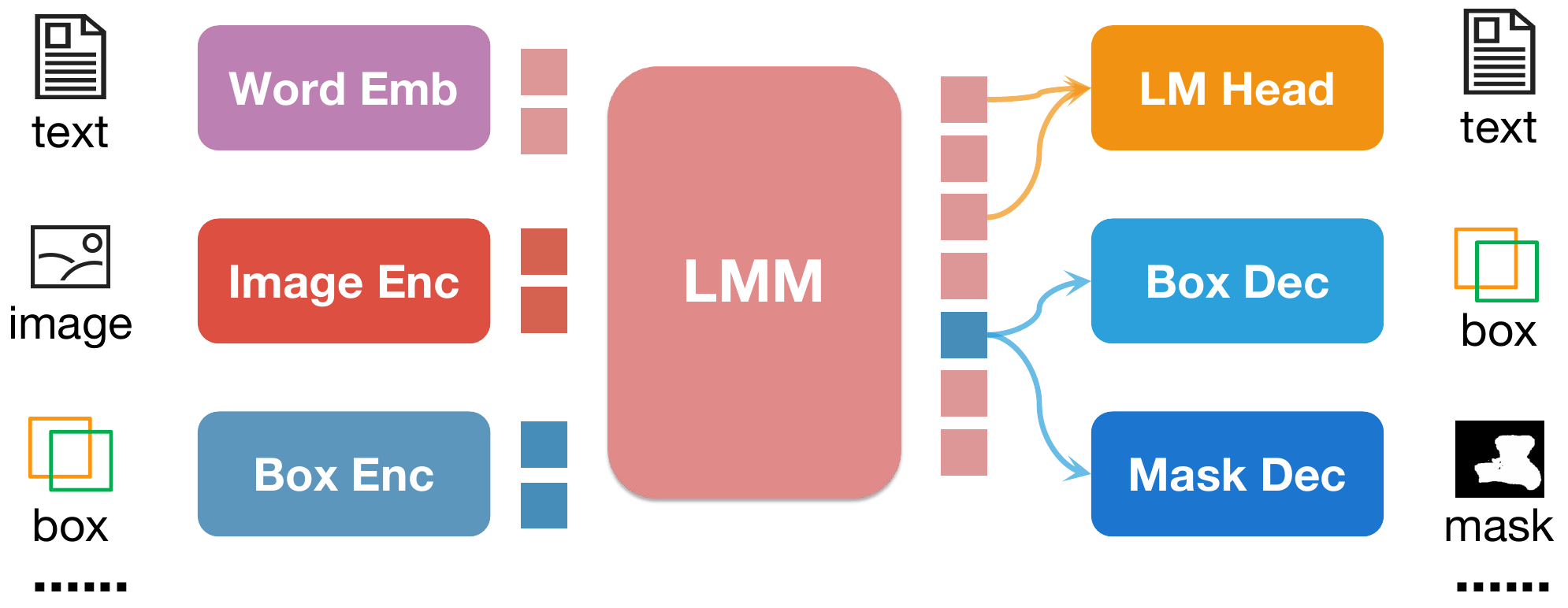}
    \captionof{figure}{By using the embedding based location modeling method, our NExT-Chat can take bounding boxes as input and output text, bounding boxes and masks in the multimodal conversation.}
    \label{fig:teaser}
\end{center}%
}]
\def\thefootnote{$*$}\footnotetext{Corresponding author.}\def\thefootnote{\arabic{footnote}}

\begin{abstract}
The development of large language models (LLMs) has greatly advanced the field of multimodal understanding, leading to the emergence of large multimodal models (LMMs). In order to enhance the level of visual comprehension, recent studies have equipped LMMs with region-level understanding capabilities by representing object bounding box coordinates as a series of text sequences (pix2seq). In this paper, we introduce a novel paradigm for object location modeling called pix2emb method, where we ask the LMM to output the location embeddings and then decode them with different decoders. 
This paradigm allows us to use different location formats (such as bounding boxes and masks) in multimodal conversations.
Leveraging the proposed pix2emb method, we train an LMM named NExT-Chat and demonstrate its capability of handling multiple tasks like visual grounding, region captioning, and grounded reasoning.
Comprehensive experiments show the effectiveness of our NExT-Chat on various tasks, e.g., NExT-Chat (87.7) vs. Shikra (86.9) on POPE-Random, NExT-Chat (68.9) vs. LISA (67.9) on referring expression segmentation task, and NExT-Chat (79.6) vs. Kosmos-2 (62.3) on region caption task.
\end{abstract}    
\section{Introduction}
\label{sec:intro}
Recently, large language models (LLMs) have shown spreading influence in different areas, among which large multimodal models (LMMs) is one of the most attractive area.
Researchers try to equip LLMs with visual perception modules resulting in LMMs~\citep{huang2023kosmos1, zhu2023minigpt4, zhang2023vpgtrans, li2023blip2} that can describe the visual content and answer visual questions.
However, these LMMs are limited to holistic image understanding without the ability to conduct region-level reasoning, for example, locating the referred objects in the conversation.

To enable region-level understanding, current solutions~\citep{peng2023kosmos2,wang2023visionllm,chen2023shikra} utilize the pix2seq~\citep{chen2021pix2seq} paradigm where the object coordinates are converted to LLM understandable text tokens (\eg, $[x_1, y_1, x_2, y_2]$).
Consequently, LMMs can output object coordinates as part of a normal next token prediction problem.
However, the pix2seq paradigm is limited to discrete coordinate outputs and struggles to provide other fine-grained formats, such as segmentation masks.

To address these limitations, we propose the pix2emb paradigm, which can accommodate different location formats.
The key idea is to model all location information as embeddings, which can be decoded into the target formats by corresponding decoders.
Specifically, we introduce two new tokens, \texttt{<trigger>} and \texttt{<loc>}, where the \texttt{<trigger>} serve as a trigger for localization and \texttt{<loc>} act as a placeholder for objects' location embeddings.
During the text generation, the \texttt{<trigger>} triggers the location decoding, where the hidden states of \texttt{<trigger>} can be used for both detection and segmentation, as depicted in Fig.~\ref{fig:method}.
Then, the predicted or provided object location will be encoded into the embedding of the \texttt{<loc>} token for object referring.
In addition to supporting flexible output formats, the pix2emb modeling also allows for the use of existing localization practices.
While the pix2seq paradigm can only frame the detection task as a token classification problem, the embedding-based paradigm formulates the localization task as a regression problem, enabling the adoption of established practices such as L1 loss, IoU loss and GIoU loss.

Building upon the proposed pix2emb method, we introduce a new LMM named NExT-Chat.
NExT-Chat is designed to handle various conversation scenarios, including visual grounding (Fig.~\ref{fig:demo_grd}), region caption (Fig.~\ref{fig:demo_region_cap}), and grounded image caption (Fig.~\ref{fig:demo_grd_cap}).
Thanks to the incorporation of LLM, NExT-Chat is also capable of handling scenarios that requires grounded reasoning.
By providing an extensive array of examples, we effectively demonstrate NExT-Chat's remarkable proficiency in understanding various components, including background elements, minute objects, and associating the objects with related knowledge.
Moreover, we validate our NExT-Chat on various datasets.
On the POPE-Random dataset, NExT-Chat achieves an impressive accuracy of 87.7, surpassing Shikra's 86.9. In referring expression segmentation (RES), it attains an average cIoU of 68.9, outperforming LISA's 67.9. Moreover, NExT-Chat achieves a remarkable 79.6 in CIDEr score for RefCOCOg region captioning, significantly exceeding Kosmos-2's 62.3.

To summarize, our contributions can be listed as follows:
\begin{itemize}[leftmargin=7.5mm]
\setlength{\itemsep}{2pt}
\item  {\it Effective Method}. We propose the pix2emb method, which can accommodate different output formats such as bounding boxes and segmentation masks.
\item {\it NExT-Chat Model}. Based on the proposed pix2emb method, we build NExT-Chat that can unify the chat, region input, detection and segmentation in a single LMM.
\item {\it Experiments and Demos}. We provide abundant qualitative and quantitative results to showcase the effectiveness of our proposed method.
\end{itemize}

\section{Related Works}
\subsection{LMM}
Large multimodal models (LMMs) are typically built on large language models (LLMs) and equipped with visual perception modules to enable the multimodal perception ability, 
which can generate captions or answer questions based on the given multimodal content.
Flamingo~\citep{alayrac2022flamingo} tries to extract vision information by a pre-trained vision backbone with a resampler, and incorporate them into the text features with a cross-attention mechanism.
Instead of using cross-attention layers, BLIP-2~\citep{li2023blip2} and Kosmos~\citep{huang2023kosmos1} directly feed the visual features into the LLMs as soft prompts.
Following BLIP-2, MiniGPT-4~\citep{zhu2023minigpt4} and VPGTrans~\citep{zhang2023vpgtrans} build LMMs with transfer learning, and significantly reduce the training cost.
For example, VPGTrans can use only around 10\% GPU hours with non-degenerated performances compared with training a new LMM from scratch.
When considering the training paradigm, researchers find that a small scale instruction tuning can better align the LMM with the expected output format.
MiniGPT-4~\citep{zhu2023minigpt4} fine-tunes its model with less than 5,000 self-instruct image-text pairs and turns the model into better conversation robot.
Different from MiniGPT-4's self-instruct, LLaVA~\citep{liu2023llava} generate the instruction tuning data with the text-only GPT-4 models by feeding the visual information as text sentences.
Otter~\citep{li2023otter, li2023mimicit} further propose a MIMIC-IT dataset that can turn the LMM into better in-context learners.
LLaVA-1.5 proposes to further fine-tune the model on human annotated datasets, which can alleviate the image-level hallucination~\citep{liu2023hallusionbench}.
However, these LMMs~\citep{alayrac2022flamingo, liu2023aligning, liu2023llava} can only take the whole image/video as input and output text, and are incapable of handling region understanding tasks.

\begin{figure*}
     \centering
     \includegraphics[width=0.8\textwidth]{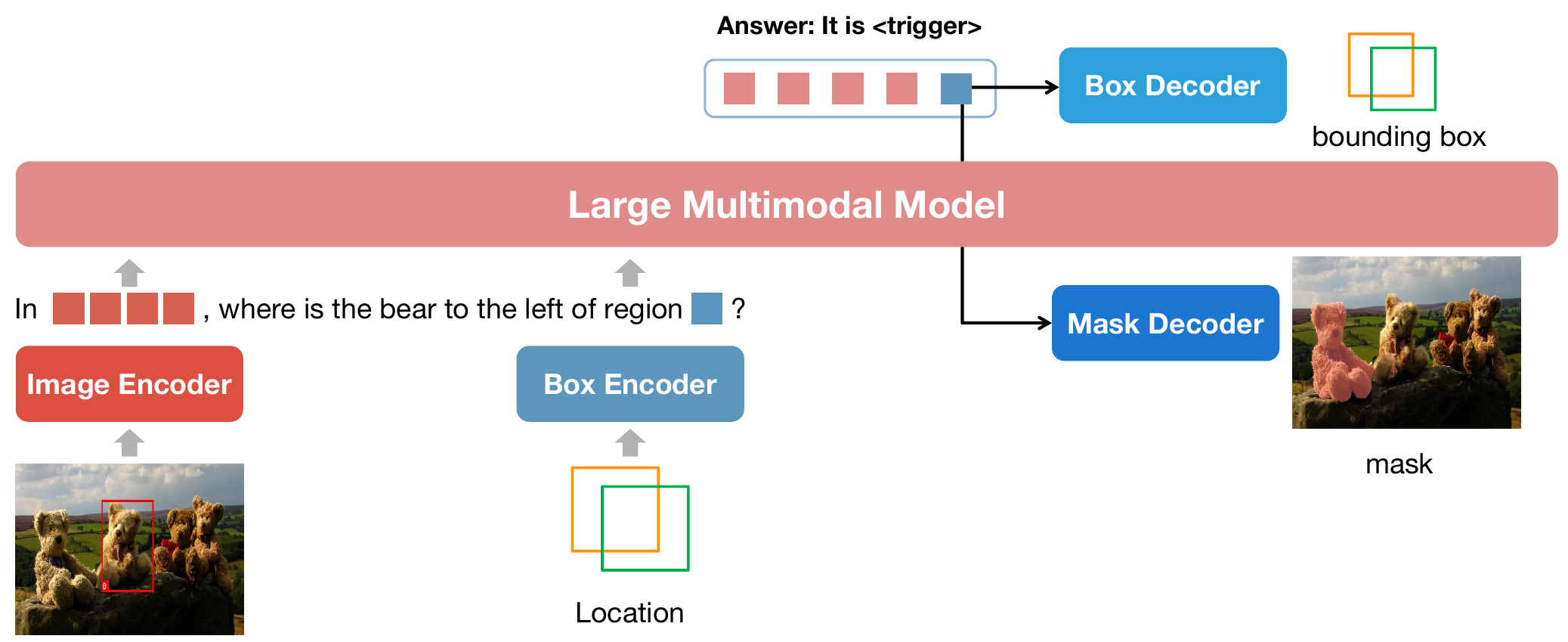}
     \caption{The overall framework of NExT-Chat.
     The image and given bounding boxes are encoded by image and box encoders respectively.
     During decoding, the hidden states of the \texttt{<trigger>} are fed into box and mask decoders, enabling object detection and segmentation.}
     \label{fig:method}
\end{figure*}

\subsection{LMM for Region Reasoning}
GPT4ROI~\citep{zhang2023gpt4roi} proposes to encode the regions as features and thus can accept the region as input.
Pix2seq~\citep{chen2021pix2seq} first propose to represent object bounding box coordinates as text tokens and thus the language model can output the object locations in a token classification manner.
However, pix2seq only validate its idea on traditional object detection tasks.
UniTab~\citep{yang2022unitab} and PEVL~\citep{yao2022pevl} further extend the idea to vision\&language tasks like visual grounding~\citep{yu2016refcoco, mao2016refcocog}.
Following this line, Vision-LLM~\citep{wang2023visionllm} and Kosmos-2~\citep{peng2023kosmos2} recently applies the token classification concept to LMMs.
Take Kosmos-2 as an example, it discretize the whole image into 32$\times$32 bins, which can be used to represent the points lying in it.
Additional 32$\times$32 tokens are introduced to the LLM's vocabulary for either coordinates input or output.
Thus, the LMM can achieve the region-level reasoning.
Shikra~\citep{chen2023shikra} point out that introducing too much new tokens will inevitably increase the training difficulties.
Thus, Shikra propose to reuse the LLM's original vocabulary and turn the box coordinates into normalized numerical values with certain precision like $[0.111, 0.111, 0.333, 0.333]$.
Although avoiding introducing too much new tokens, it requries roughly 26 tokens to represent each bounding box, which is ineffective.
Different from these works, we do not formulate the object localization problem as a token classification problem.
Our NExT-Chat introduces an \texttt{<trigger>} token as the trigger for location decoding, and then use the hidden states to decode the bounding boxes and the segmentation masks.

\section{Method}


In this section, we present the NExT-Chat framework, starting with an introduction to the overall LMM architecture ($\S$\ref{sec:arch}), followed by a description of the pix2emb method ($\S$\ref{sec:method_emb}). Additionally, we provide details on the training process ($\S$\ref{sec:method_training}).

\subsection{LMM Architecture}
\label{sec:arch}
For the LMM architecture, we adopt a LLaVA-like architecture.
Specifically, we employ a CLIP ViT-L/14@336px~\citep{radford2021clip} as the vision encoder.
The input image is converted into 24$\times$24 patch embeddings and then projected to the same dimension as the word embeddings of the LLM.
These patch embeddings serve as visual tokens.
Then, the visual tokens will be fed into a decoder-only LLM for conditional text generation.
Regarding the selection of LLMs, we opt for the recently released Vicuna-1.5 model~\citep{zheng2023vicuna}.

\subsection{Pix2Emb Method}
\label{sec:method_emb}
\textbf{Detection.}
To model the object location as output, we introduce a special token, denoted as \texttt{<trigger>}, which serves to trigger the localization.
As depicted in Fig.~\ref{fig:method}, the LMM is trained to generate the \texttt{<trigger>} token before predicting the locations.
Then, the embedding $\mathbf{t} \in \mathcal{R}^n$ of \texttt{<trigger>} is then passed to the \textit{Box Decoder} $\mathcal{F}$ for regression.
Mathematically, this can be expressed as follows:
\begin{equation}
    \mathbf{b} = \mathcal{F}(\mathbf{t}),
\end{equation}
where $\mathbf{b} \in \mathcal{R}^4$ represents the predicted bounding box coordinates in the format $[x_0, y_0, x_1, y_1]$.


In our NExT-Chat model, the box decoder consists of a 2-layer MLP.
To supervise the location output, we employ a joint loss function comprising of the L1 loss and the GIoU loss~\citep{rezatofighi2019giou} during training:
\begin{equation}
    {\cal L}_{det} = \alpha{\cal L}_1(\mathbf{b}, \mathbf{b}_{gt})+\beta\text{GIoU}(\mathbf{b}, \mathbf{b}_{gt}),
\end{equation}
where $\mathbf{b}_{gt}$ represents the ground truth coordinates, and $\alpha=2$, $\beta=0.8$ follows the ratio utilized in DETR~\citep{carion2020detr}.

\smallskip
\noindent
\textbf{Segmentation.}
Similar to the detection process, we utilize the hidden states $\mathbf{t}$ of the \texttt{<trigger>} as input for the mask head.
Inspired by LISA~\citep{lai2023lisa}, we use SAM~\citep{kirillov2023sam} as our mask head, which also additionally takes the original image as input.
To ensure compatibility between the hidden states and SAM, we first project the hidden states to match the dimension of SAM's prompt embedding using a linear projector. Subsequently, the projected hidden states are fed as the prompt embedding to SAM.
For improved performance, we also encode the detected bounding boxes into a prompt embedding with SAM's prompt encoder and concatenate it with the projected embedding.
To train the mask output, we follow the practice outlined in lightning-SAM\footnote{\url{https://github.com/luca-medeiros/lightning-sam/tree/main}}:
\begin{equation}
    {\cal L}_{seg} = \text{IoU}(\mathbf{m}, \mathbf{m}_{gt})+\text{D}(\mathbf{m}, \mathbf{m}_{gt}) + \beta \text{F}(\mathbf{m}, \mathbf{m}_{gt}),
\end{equation}
where $\text{IoU}$, $\text{D}$, and $\text{F}$ are IoU Loss, Dice Loss, and Focal Loss separately.
$\beta$ is set to 20 in our experiments.

\begin{figure}
     \centering
     \includegraphics[width=0.9\linewidth]{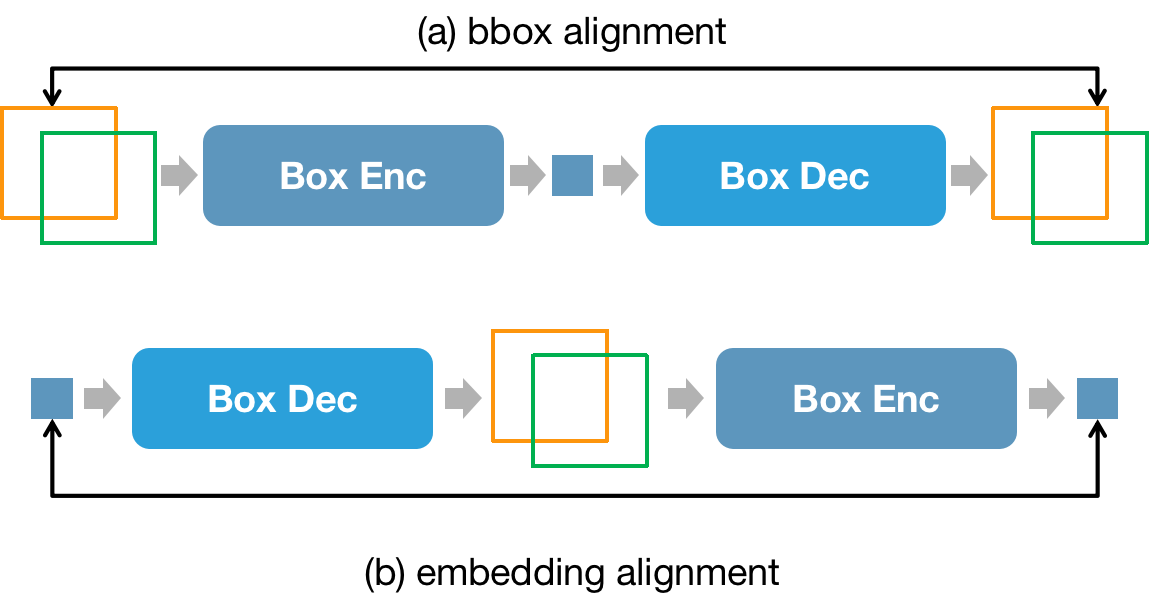}
     \caption{Cycle loss utilized to bind box encoder and decoder training.}
     \label{fig:cyc_loss}
      \vspace{-4mm}
 \end{figure}

\smallskip
\noindent
\textbf{Location as Input.}
In addition to the location output, it is essential to incorporate location as input as well.
To be consistent with the location output modeling, we also use a single embedding to represent the location information.
Therefore, the output location embedding can also serve as the input embedding.
Consequently, we introduce another 2-layer MLP, referred to as the location encoder $\mathcal{G}$.
In order to simplify the problem, we convert all location formats into bounding boxes $b$ and subsequently transform them into embeddings $\mathbf{t} \in \mathbb{R}^n$ suitable for the LLM.
The location encoder can be supervised through the standard text generation loss ${\cal L}_{text}$. For instance, when inquiring about the relationship between bounding box $\mathbf{b}_1$ and $\mathbf{b}_2$, the location encoder is compelled to provide precise information.

However, we observe that the location encoder cannot be effectively trained solely through indirect supervision from ${\cal L}_{text}$.
As a result, we introduce an additional cycle loss to facilitate the training of the encoder in conjunction with the decoder.
As illustrated in Fig.~\ref{fig:cyc_loss} (a), a bounding box will be encoded and then decoded, where two bounding boxes are asked to be the same.
Similarly, the hidden states of \texttt{<trigger>} will also be used to calculate the cycle loss (Fig.~\ref{fig:cyc_loss} (b)).
Formally, the $L_{cyc}$ is defined as:
\begin{equation}
    {\cal L}_{cyc} = {\cal L}_1(\mathbf{b}, \mathcal{F}(\mathcal{G}(\mathbf{b}))) + {\cal L}_2(\mathbf{t}, \mathcal{G}(\mathcal{F}(\mathbf{t}))),
\end{equation}
where $\mathbf{b}$ and $\mathbf{t}$ are provided bounding box and predicted embedding respectively.
Additionally, ${\cal L}_1$ and ${\cal L}_2$ correspond to the L1 Loss and L2 Loss, respectively.



\subsection{Training Process}
\label{sec:method_training}
We employ a three-stage training process, consisting of pre-training, instruction tuning, and segmentation training, to train our model.
The idea is to train the bounding box decoding ability for the first two stages and then extend to segmentation with a lightweight training.

\smallskip
\noindent
\textbf{Stage 1.}
During this stage, we perform pre-training using a mixture of data from various sources, including Flickr30K Entities~\citep{plummer2015flickr30k}, Visual Genome~\citep{krishna2017vg}, RefCOCO~\citep{yu2016refcoco}, RefCOCO+~\citep{yu2016refcoco}, RefCOCOg~\citep{mao2016refcocog}, VQAv2~\citep{antol2015vqa}, PointQA~\citep{mani2020pointqa}, Visual7W~\citep{zhu2016visual7w}, and VCR~\citep{zellers2019vcr}.
The model is trained with a batch size of 64 and a learning rate of 2e-5 for 65k steps. During this pre-training stage, the entire language model with the box decoder, is trained while keeping the image encoder frozen.
The training loss is formulated as:
\begin{equation}
    {\cal L}_{s1} = {\cal L}_{text} + {\cal L}_{det} + {\cal L}_{cyc}.
\end{equation}
For NExT-Chat 7B model, the stage-1 training uses 8 A100 (80G) GPUs for around 59 hours.

\smallskip
\noindent
\textbf{Stage 2.}
In the second stage, we further fine-tune the model using data from VQAv2, RefCOCO, Flickr30K Entities, LLaVA-instruct, VG grounded captioning, VCR, and Shikra-RD~\citep{chen2023shikra}. The batch size is reduced to 64, and the learning rate is set to 2e-5.
The loss is the same with stage-1's loss.
For NExT-Chat 7B model, the stage-2 training uses 8 A100 (80G) GPUs for around 10 hours.

\smallskip
\noindent
\textbf{Stage 3.}
After the two stages training, the model is equipped with the ability to engage in dialogue and perform image localization.
To prevent catastrophic forgetting, we keep most of the parameters frozen during the segmentation training.
Specifically, we only train the linear projector between the LMM and SAM, as well as the decoder of SAM.
The loss for the stage-3 is:
\begin{equation}
    {\cal L}_{s3} = {\cal L}_{seg}.
\end{equation}
Thanks to the small amount of training parameters, the training can be done in 3 hours with 8 A100 (80G) GPUs.
This training is performed using the referring segmentation splits of RefCOCO, RefCOCO+, and RefCOCOg datasets.

\section{Experiment}

\begin{table*}
\centering
\caption{\textbf{Image Hallucination:} the comparison between our NExT-Chat with current SOTA models on the POPE benchmark for image hallucination diagnosis.
}
\label{tab:pope_results}
\resizebox{\textwidth}{!}{%
\begin{tabular}{l|l|>{\columncolor[RGB]{230, 242, 255}}ccccccc}
\toprule
Datasets & Metrics & NExT-Chat & Shikra & InstructBLIP  & MiniGPT-4 & LLaVA &MM-GPT & mPLUG-Owl \\
\cmidrule(lr){1-2}\cmidrule(lr){3-9}
\multirow{5}{*}{Random}
& Accuracy ($\uparrow$)      & 87.70 & 86.90 & 88.57 & 79.67 &50.37  & 50.10& 53.97 \\
& Precision ($\uparrow$)     & 93.46 & 94.40 & 84.09 &78.24  &50.19  & 50.05&52.07\\
& Recall ($\uparrow$)       & 81.87  & 79.27  & 95.13 &82.20  &  99.13& 100.00&99.60  \\
& F1-Score ($\uparrow$)      & 87.28 & 86.19 &89.27  &80.17  & 66.64 &  66.71 &68.39 \\
& Yes  & 45.15  & 43.26  & 56.57 & 52.53 & 98.77 &  99.90&95.63 \\
\cmidrule(lr){1-2}\cmidrule(lr){3-9}
\multirow{5}{*}{Popular}
& Accuracy ($\uparrow$)     &84.57  & 83.97  & 82.77 &69.73  &49.87  & 50.00&50.90  \\
& Precision ($\uparrow$)    &86.54  & 87.55 & 76.27 & 65.86 &49.93  & 50.00&50.46  \\
& Recall ($\uparrow$)       &81.87  & 79.20  & 95.13 &81.93  & 99.27 & 100.00&99.40  \\
& F1-Score ($\uparrow$)     &84.14  & 83.16 & 84.66 & 73.02 & 66.44 & 66.67 & 66.94\\
& Yes   & 47.30 & 45.23  & 62.37 & 62.20 & 99.40 &100.00  &98.57\\
\cmidrule(lr){1-2}\cmidrule(lr){3-9}
\multirow{5}{*}{Adversarial}
& Accuracy ($\uparrow$)      & 81.93 & 83.10  & 72.10  &65.17  &  49.70& 50.00 & 50.67\\
& Precision ($\uparrow$)     & 82.02 & 85.60  & 65.13 & 61.19 & 49.85 & 50.00 & 50.34\\
& Recall ($\uparrow$)       & 81.80  & 79.60 & 95.13 & 82.93 & 99.07 & 100.00 & 99.33\\
& F1-Score ($\uparrow$)     & 81.91  & 82.49 & 77.32 &  70.42& 66.32 &66.67  & 66.82\\
& Yes   & 49.87 & 46.50 & 73.03 &67.77  & 99.37 &   100.00&98.67\\
\bottomrule
\end{tabular}%
}
\end{table*}

\begin{table*}
    \centering
    \caption{\textbf{RES:} comparison between our NExT-Chat and baselines on RES. The evaluation metric is \textbf{cIoU}.}
    \resizebox{0.7\textwidth}{!}{
    \begin{tabular}{l c c c c c c c c}
    \toprule
    \multirow{2}{*}{Methods} & \multicolumn{3}{c}{RefCOCO} & \multicolumn{3}{c}{RefCOCO+} & \multicolumn{2}{c}{RefCOCOg} \\
     \cmidrule(lr){2-4} \cmidrule(lr){5-7} \cmidrule(lr){8-9} 
     & val & testA & testB & val & testA & testB & val & test \\
    \midrule
    MCN~\citep{luo2020mcn} & 62.4 & 64.2 & 59.7 & 50.6 & 55.0 & 44.7 & 49.2 & 49.4 \\
    VLT~\citep{ding2021vlt} & 67.5 & 70.5 & 65.2 & 56.3 & 61.0 & 50.1 & 55.0 & 57.7 \\
    CRIS~\citep{wang2022cris} & 70.5 & 73.2 & 66.1 & 65.3 & 68.1 & 53.7 & 59.9 & 60.4 \\
    LAVT~\citep{yang2022lavt} & 72.7 & 75.8 & 68.8 & 62.1 & 68.4 & 55.1 & 61.2 & 62.1 \\
    GRES~\citep{liu2023gres} & 73.8 & 76.5 & 70.2 & 66.0 & 71.0 & 57.7 & 65.0 & 66.0 \\
    X-Decoder~\citep{zou2023xdecoder} & - & - & - & - & - & - & 64.6 & - \\
    SEEM~\citep{zou2023seem} & - & - & - & - & - & - & 65.7 & - \\
    LISA-7B~\citep{lai2023lisa} & 74.1 & 76.5 & 71.1 & 62.4 & 67.4 & 56.5 & 66.4 & 68.5 \\
    \midrule
    \rowcolor[RGB]{230, 242, 255} NExT-Chat (\textbf{ours}) & 74.7 & 78.9 & 69.5 & 65.1 & 71.9 & 56.7 & 67.0 & 67.0 \\
    \bottomrule
    \end{tabular}}
    \label{tab:res}
\end{table*}

\begin{table*}
    \centering
    \caption{\textbf{REC:} comparison between our NExT-Chat and baselines on REC. The evaluation metric is \textbf{Acc@0.5}. * refers to the specialist or fine-tuned methods.} 
    \resizebox{1\textwidth}{!}{
    \begin{tabular}{l l c c c c c c c c}
    \toprule
    \multirow{2}{*}{Type} & \multirow{2}{*}{Methods} & \multicolumn{3}{c}{RefCOCO} & \multicolumn{3}{c}{RefCOCO+} & \multicolumn{2}{c}{RefCOCOg} \\
     \cmidrule(lr){3-5} \cmidrule(lr){6-8} \cmidrule(lr){9-10} 
    & & val & testA & testB & val & testA & testB & val & test \\
    \midrule
    \multirow{7}{*}{non-LMM}
    & MAttNet* ~\citep{yu2018mattnet} & 76.4 & 80.4 & 69.3 & 64.9 & 70.3 & 56.0 & 66.7 & 67.0 \\
    & OFA-L~\citep{wang2022ofa} & 80.0 & 83.7 & 76.4 & 68.3 & 76.0 & 61.8 & 67.6 & 67.6  \\
    & OFASys & - & 80.1 & - & - & - & - & - & -  \\
    & TransVG* ~\citep{deng2021transvg} & 81.0 & 82.7 & 78.4 & 64.8 & 70.7 & 56.9 & 68.7 & 67.7 \\
    & UNITER* ~\citep{chen2020uniter} &  81.4 & 87.0 & 74.2 & 75.9 & 81.5 & 66.7 & 74.0 & 68.7 \\
    & VILLA* ~\citep{gan2020villa} &  82.4 & 87.5 & 74.8 & 76.2 & 81.5 & 66.8 & 76.2 & 76.7 \\
    & UniTAB* ~\citep{yang2022unitab} &  86.3 & 88.8 & 80.6 & 78.7 & 83.2 & 69.5 & 80.0 & 80.0 \\
    & G-DINO-L* ~\citep{liu2023gdino} &  90.6 & 93.2 & 88.2 & 82.8 & 89.0 & 75.9 & 86.1 & 87.0 \\
    \midrule
    \multirow{3}{*}{LMM (pix2seq)}
    & VisionLLM-H~\citep{wang2023visionllm} & - & 86.7 & - & - & - & - & - & -  \\
    & Shikra-7B~\citep{chen2023shikra} & 87.0 & 90.6 & 80.2 & 81.6 & 87.4 & 72.1 & 82.3 & 82.2 \\
    & Shikra-13B~\citep{chen2023shikra} & 87.8 & 91.1 & 81.8 & 82.9 & 87.8 & 74.4 & 82.6 & 83.2 \\
    \midrule
    \rowcolor[RGB]{230, 242, 255} \multirow{1}{*}{LMM (pix2emb)} & NExT-Chat-7B (\textbf{ours}) & 85.5 & 90.0 & 77.9 & 77.2 & 84.5 & 68.0 & 80.1 & 79.8 \\
    \bottomrule
    \end{tabular}}
    \label{tab:rec}
\end{table*}

\begin{table}
    \caption{\textbf{Region Captioning}: comparison between our NExT-Chat and baselines on RefCOCOg.}
    \centering
    \begin{tabular}{l c c}
    \toprule
    \multirow{2}{*}{Methods} & \multicolumn{2}{c}{RefCOCOg} \\
    \cmidrule(lr){2-3}
    & CIDEr& METEOR \\
    \midrule
    GRIT~\citep{wu2022grit} & 71.6 & \textbf{15.2} \\
    Kosmos-2~\citep{peng2023kosmos2} (0-shot) & 60.3  & 12.2 \\
    Kosmos-2~\citep{peng2023kosmos2} (2-shot) &  62.2 & 13.8 \\
    Kosmos-2~\citep{peng2023kosmos2} (4-shot) &  62.3 & 14.1  \\
    ASM~\citep{wang2023allseeing} & 41.9 & 13.6 \\
    \midrule
    \rowcolor[RGB]{230, 242, 255} NExT-Chat (\textbf{ours}) & \textbf{79.6}  & 12.0 \\
    \bottomrule
    \vspace{-10mm}
    \end{tabular}
    \label{tab:reg_cap}
\end{table}

In this section, we begin by conducting a rigorous evaluation to validate the effectiveness of our pix2emb approach in a fair comparison setting.
Following that, we demonstrate the potential of our NExT-Chat model by presenting a wide range of qualitative results from different scenarios.
Finally, we provide quantitative results to compare the performance of our NExT-Chat model with the current SOTA methods on the image-level hallucination, referring expression segmentation, referring expression detection and region-level caption tasks.

\subsection{Applications across Different Scenarios}
\label{sec:exp_demp}
In this section, we present qualitative results that showcase the capabilities of our NExT-Chat model across various scenarios.

\begin{figure*}[h]
     \centering
     \includegraphics[width=1\textwidth]{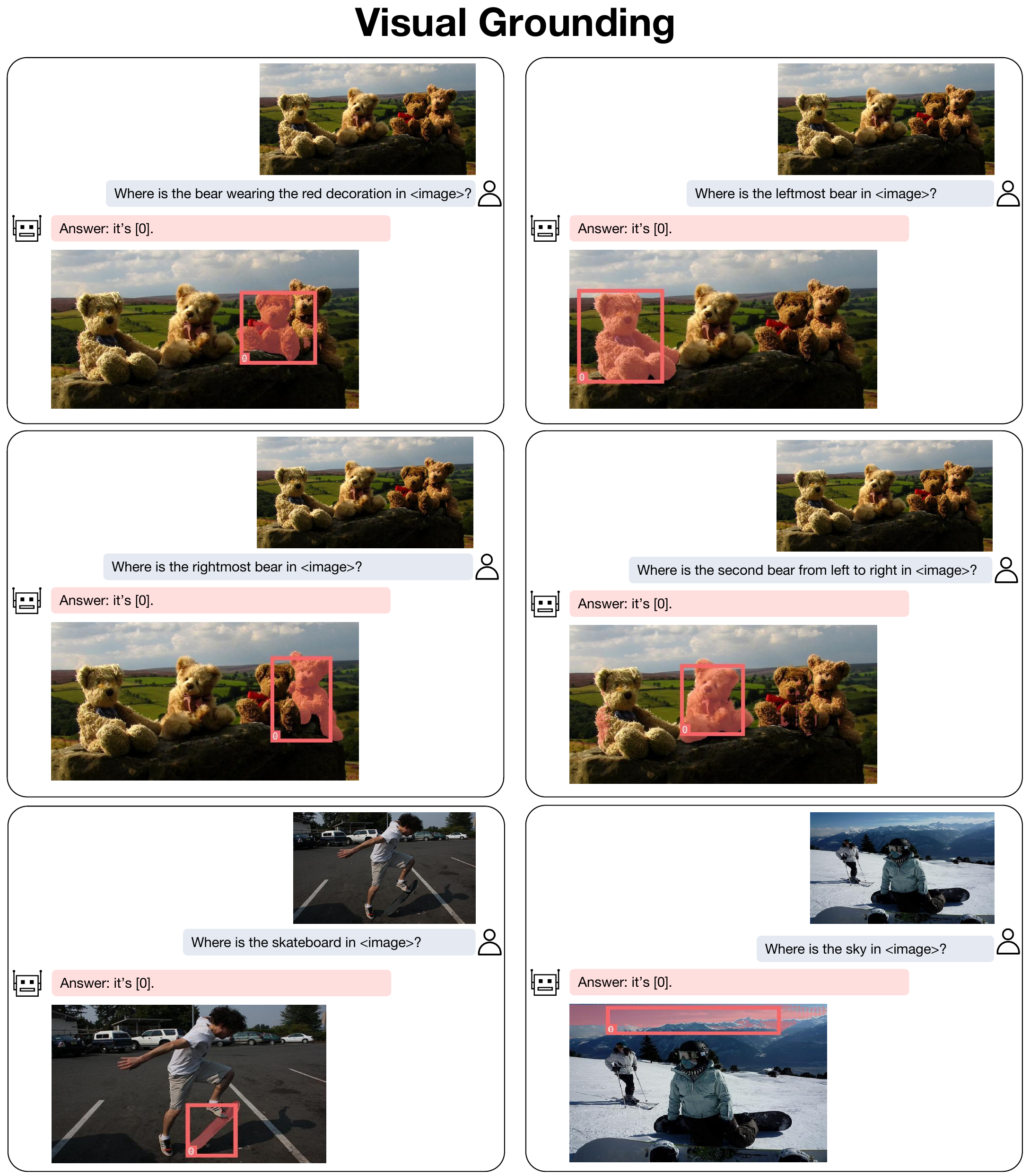}
     \caption{Visual grounding examples of NExT-Chat.}
     \label{fig:demo_grd}
 \end{figure*}

\begin{figure*}
     \centering
     \includegraphics[width=1\textwidth]{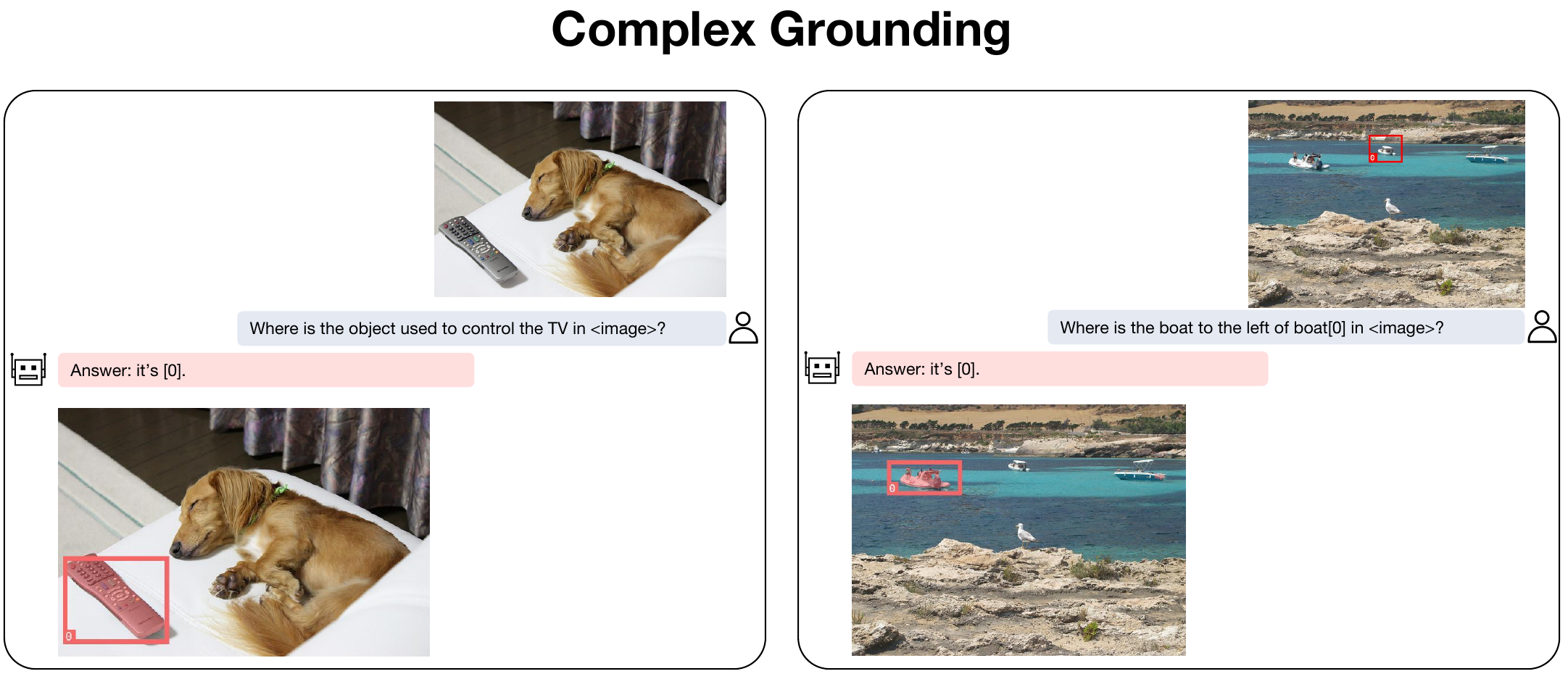}
     \caption{Hard visual grounding examples of NExT-Chat.}
     \label{fig:demo_grd_cplex}
 \end{figure*}

\smallskip
\noindent
\textbf{Visual Grounding.}
As shown in Fig.~\ref{fig:demo_grd}, we can see that our NExT-Chat accurately detects and segments the queried objects, such as the bears and the sky in the background.
To ensure that our model is not biased towards specific objects, we test it with different queries to find all four bears individually.
Our model successfully localizes each bear based on the given queries.
Additionally, our model showcases reasoning abilities through challenging grounding problems.
For instance, in Fig.~\ref{fig:demo_grd_cplex}, our model accurately localizes the remote in response to the query ``Where is the object to control the TV in image?" It also localizes the boat based on understanding the given object location input.

\begin{figure*}[h]
     \centering
     \includegraphics[width=1\textwidth]{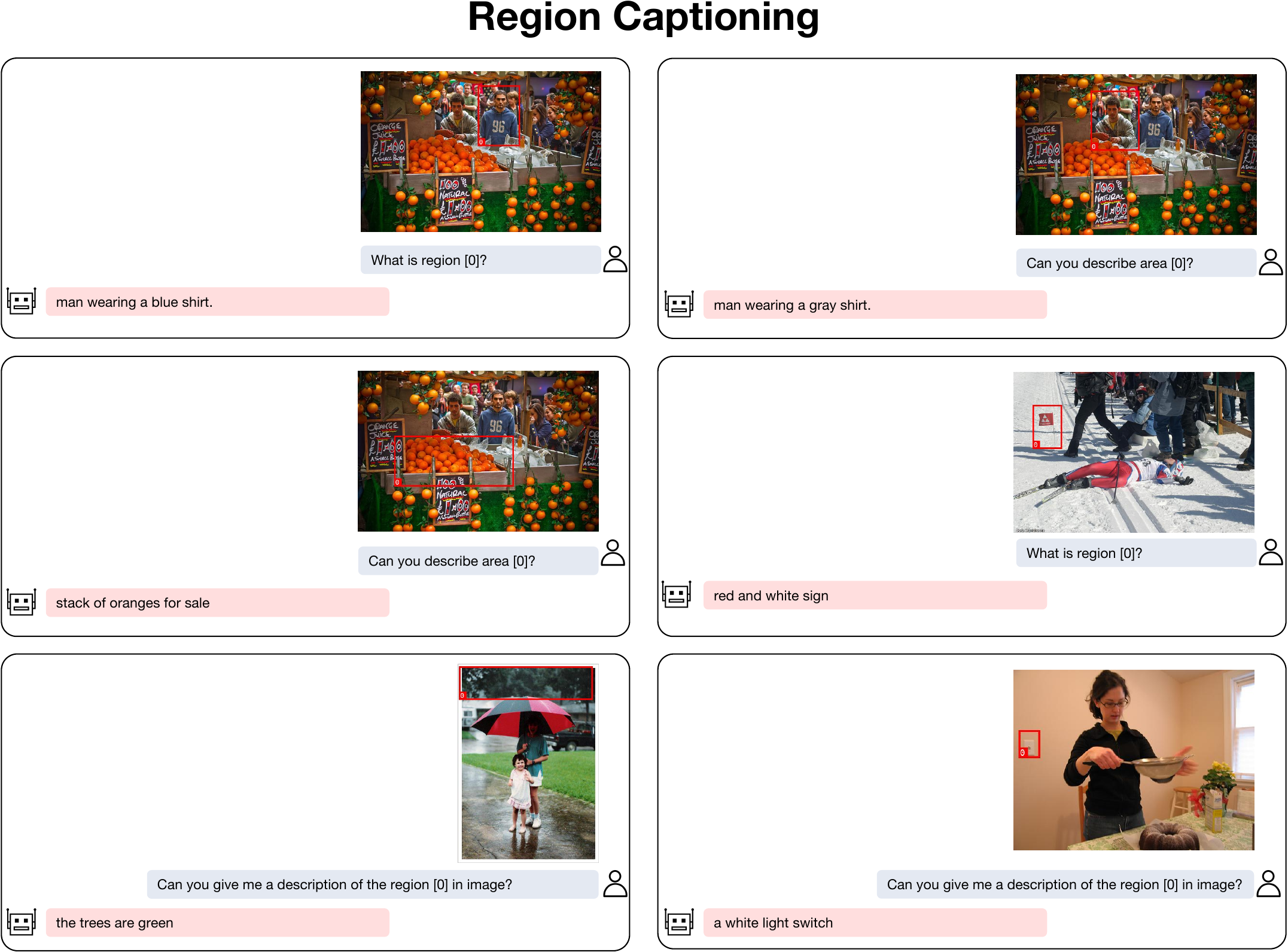}
     \caption{Region captioning examples of NExT-Chat.}
     \label{fig:demo_region_cap}
 \end{figure*}

\smallskip
\noindent
\textbf{Region Captioning.}
To evaluate the effectiveness of our NExT-Chat model for region input, we conducted experiments where the model generates descriptions based on given bounding boxes. 
As depicted in Fig.~\ref{fig:demo_region_cap}, our model consistently produces accurate descriptions specifically tailored to the provided regions, without being influenced by the overall image content or salient regions. 
We observed this behavior consistently across different examples. Notably, in the second row of Fig.~\ref{fig:demo_region_cap}, our model demonstrates the ability to accurately recognize and describe small objects such as flags, as well as background objects like trees.
This demonstrates the robustness and effectiveness of our model in generating region-based captions.



\begin{figure*}[h]
     \centering
     \includegraphics[width=1\textwidth]{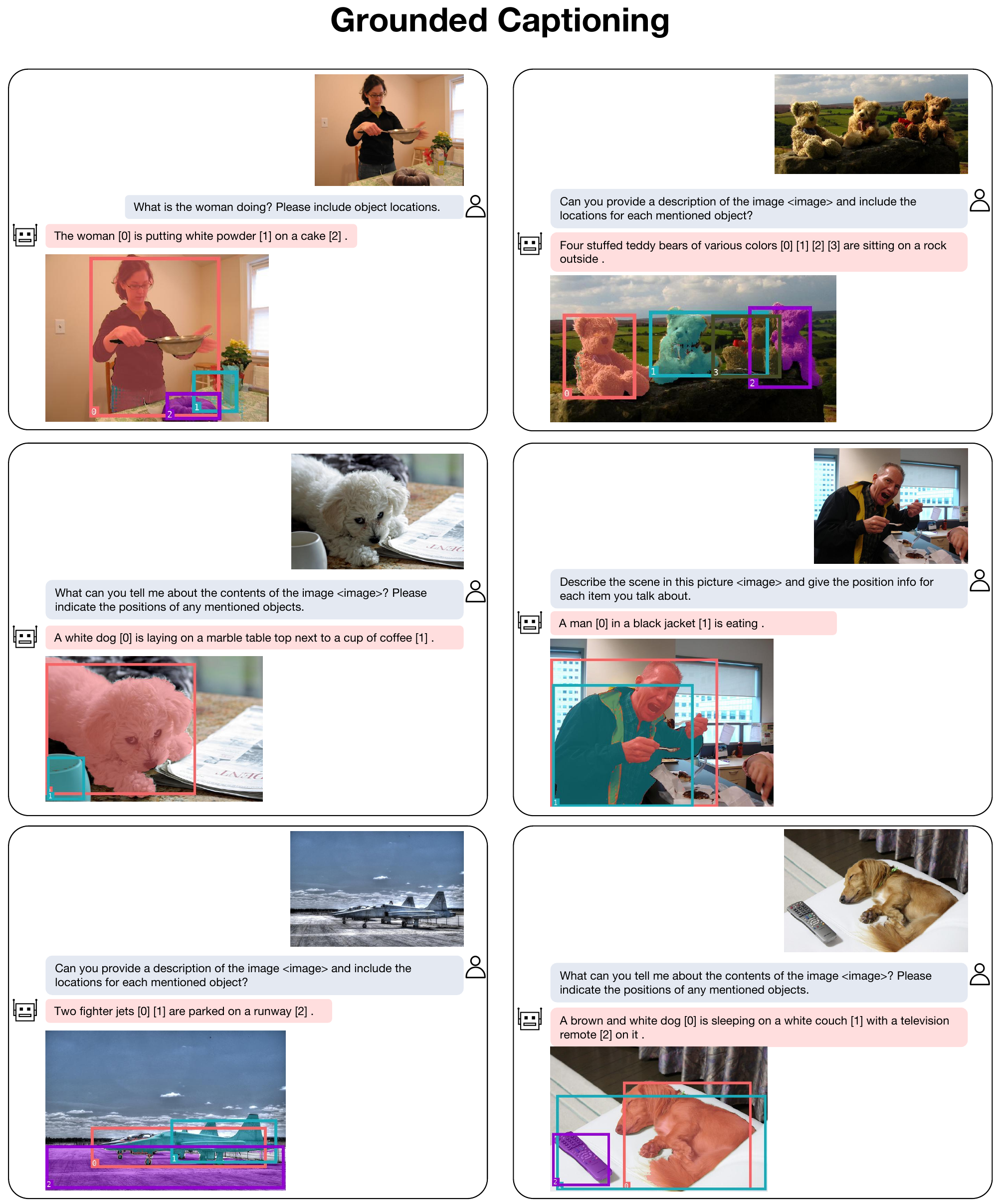}
     \caption{Grounded captioning examples of NExT-Chat.}
     \label{fig:demo_grd_cap}
 \end{figure*}

\smallskip
\noindent
\textbf{Grounded Captioning.}
Another compelling application of our NExT-Chat model is its ability to describe images by referencing specific objects present within them.
Fig.~\ref{fig:demo_grd_cap} demonstrates that our model can accurately identify and describe the major 2 or 3 objects in an image, effectively organizing them into coherent sentences.
By incorporating object references, our model demonstrates a reduced tendency to generate captions containing non-existent objects. This highlights the model's capability to generate more accurate and contextually grounded image descriptions.

\begin{figure*}[h]
     \centering
     \includegraphics[width=1\textwidth]{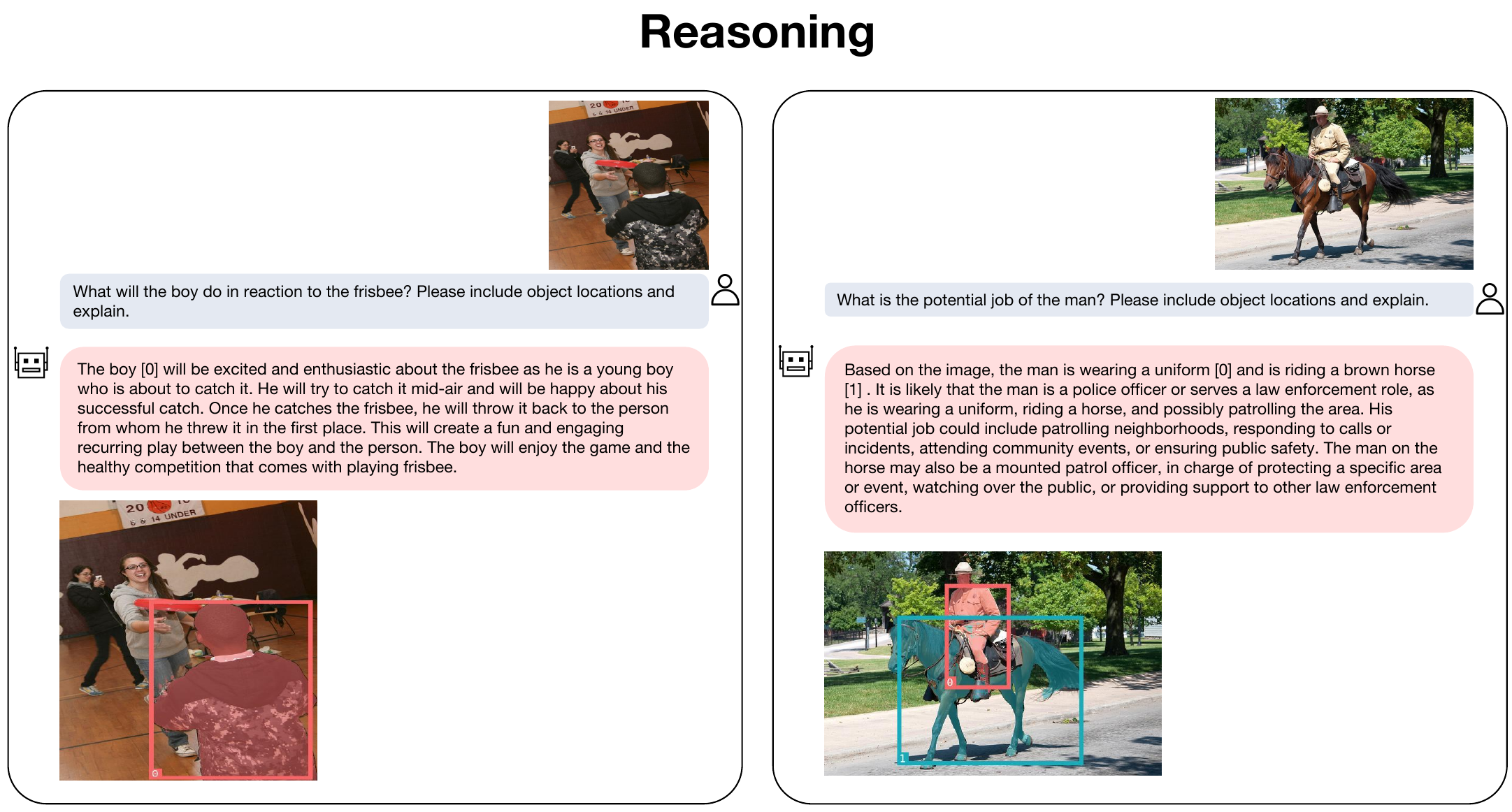}
     \caption{Reasoning examples of NExT-Chat.}
     \label{fig:demo_reason}
\end{figure*}

\smallskip
\noindent
\textbf{Reasoning.}
In addition to its demonstrated ability in single-turn and concise response generation, our NExT-Chat model also possesses the capability for generating detailed explanations in response to given questions.
As illustrated in the third example of Fig.~\ref{fig:demo_reason}, our model exhibits the ability to infer the occupation of the man in the image by analyzing contextual cues such as his uniform and the horse he is riding.
This inference is supported by the model's ability to localize relevant regions within the image.
Furthermore, for each hypothesis regarding the man's occupation, our model provides detailed descriptions of the potential duties associated with that occupation. 
This showcases the model's capacity for nuanced reasoning and comprehensive explanation generation.

\section{Comparison with SOTAs}
In this study, we evaluate our NExT-Chat model by comparing it with current state-of-the-art (SOTA) models on various tasks including image-level hallucination diagnose (POPE dataset~\citep{li2023pope}), referring detection, referring segmentation and region-level captioning (RefCOCOg).

\subsection{Hallucination}
\noindent
\smallskip
\textbf{Experimental Setup.} For a comprehensive evaluation, we benchmarked our NExT-Chat model against current state-of-the-art (SOTA) LMMs including Shikra~\citep{chen2023shikra}, InstructBLIP, MiniGPT-4~\citep{zhu2023minigpt4}, LLaVA~\citep{liu2023llava}, MM-GPT~\citep{gong2023mmgpt} and mPLUG-OWL~\citep{ye2023mplug} on the POPE dataset~\cite{li2023pope}.

\noindent
\smallskip
\textbf{Results.}
The results, presented in Table~\ref{tab:pope_results}, demonstrate that our NExT-Chat model exhibits competitive performance compared with existing SOTA models.
Notably, our model achieves the the best performance for the random and popular splits and achieve the second best performance of the adversatrial split.
These findings indicate that our NExT-Chat model is competent in generating accurate responses, thus positioning it among the top-performing models in the field.

\subsection{Referring Expression Segmentation}
\noindent
\smallskip
\textbf{Experimental Setup.} To rigorously assess our model's proficiency in generating segmentation masks guided by natural language instructions, we use the referring expression segmentation (RES) splits of RefCOCO, RefCOCO+, and RefCOCOg.
As for baselines, we choose both the LMM based method (LISA~\citep{lai2023lisa}) and non-LMM based methods including MCN~\citep{luo2020mcn}, VLT~\citep{ding2021vlt}, CRIS~\citep{wang2022cris}, LAVT~\citep{yang2022lavt}, GRES~\citep{liu2023gres}, X-Decoder~\citep{zou2023xdecoder} and SEEM~\citep{zou2023seem}.
cIoU metric is employed to evaluate different methods.

\noindent
\smallskip
\textbf{Results.}
As demonstrated in Table~\ref{tab:res}, NExT-Chat exhibits superior or comparable cIoU scores relative to all baseline models. In comparison with non-LMM based methods, our approach consistently achieves either the highest or second-highest performance across various dataset splits, with the sole exception being the RefCOCO+ val set. Against LMM-based methods, specifically the LISA-7B model, NExT-Chat demonstrates enhanced performance in six dataset splits, notably achieving a substantial 4.5-point improvement in the RefCOCO+ testA split. It is noteworthy that NExT-Chat is trained with a significantly smaller dataset, comprising only 127k object segmentation masks, in stark contrast to baselines such as LISA, which utilize datasets more than an order of magnitude larger.
These results underscore the efficiency of our training paradigm in substantially reducing the dependency on extensive and costly segmentation annotation datasets.

\subsection{Referring Expression Comprehension}
\noindent
\smallskip
\textbf{Experimental Setup.} In addition to the segmentation ability, we also validate the detection ability of our method.
Concretely, we adopt the REC splits of RefCOCO, RefCOCO+, and RefCOCOg.
As for baselines, we first include the LMM method (pix2seq): VisionLLM-H~\citep{wang2023visionllm}, and Shikra~\citep{chen2023shikra}
We also include the non-LLM based methods: MAttNet~\citep{yu2018mattnet}, OFA-L~\citep{wang2022ofa}, UniTab~\citep{yang2022unitab}, G-DINO-L~\citep{liu2023gdino} and etc, where the models with * mark in the Table~\ref{tab:rec} refer to the specialist and fine-tuned methods.

\noindent
\smallskip
\textbf{Results.}
First of all, our NExT-Chat can achieve excellent REC results and can even beat a series of fine-tuned methods like VILLA~\citep{gan2020villa}, UNITER~\citep{chen2020uniter} and TranVG~\citep{deng2021transvg} on all of the splits. 
There is also an interesting phenomenon that our NExT-Chat is slightly lower than Shikra-7B even with a similar data recipe for detection training.
We hypothesize the reasons are that: (1) it is difficult to seek a perfect balance between the LM loss and localization loss, where the pix2seq methods do not suffer from this problem. (2) LLM is not pre-trained on the regression tasks and will potentially increase the training difficulty.
However, we believe that incorporating the regression tasks in the LMM will be necessary, especially for targets like embodied AI.

\subsection{Region Caption}
\noindent
\smallskip
\textbf{Experiment Setup.} In addition to the region output, we also validate the model's ability of taking regions as input.
The RefCOCOg is adopted, where each model is asked to describe the given region.
The CIDEr and METEOR are applied as the evalution metrics.
For the baselines, we choose GRIT~\citep{wu2022grit}, Kosmos-2~\citep{peng2023kosmos2} and ASM~\citep{wang2023allseeing}.

\noindent
\smallskip
\textbf{Results.}
As shown in Fig.~\ref{tab:reg_cap}, our model is capable of achieving the best performance on CIDEr across all of the baselines, which shows superiority of our NExT-Chat.
Especially for Kosmos-2, we can beat the version with 4-shot examples.

 
 \section{Conclusion}
In this paper, we present a novel location modeling method called pixel2emb, which utilizes embeddings to achieve multiple location output formats, such as bounding boxes and segmentation masks.
Through comprehensive exploratory experiments, we demonstrate the effectiveness of the proposed pix2emb method.
Additionally, we train a LMM named NExT-Chat, which significantly broadens the range of application scenarios for LMMs.
Our NExT-Chat exhibits the ability to handle diverse tasks, including visual grounding, region captioning, grounded captioning and complex question reasoning.
In the future, we will continue to enhance the model's ability on conducting better detection and segmentation.
Another promising direction is to extend the NExT-Chat model to multimodal agent which can handle complex tasks that requires region understanding.

\section{Limitation}
In the training procedure, our dataset primarily comprises individual image inputs, resulting in a limitation of our NExT-Chat model when it comes to handling multiple image inputs. Furthermore, the absence of sufficient training data from diverse domains hinders the model's ability to generate accurate predictions in tasks involving medical and satellite image analysis.

\section*{Author Contributions}
Ao Zhang initializes the project, conducts experiments and writes the main part of the paper.
Wei Ji and Yuan Yao proof read the paper.
Tat-Seng Chua, Zhiyuan Liu and Yuan Yao provides valuable suggestions on the paper structure, experiment design and paper revision.

\clearpage

\clearpage

{
    \small
    \bibliographystyle{ieeenat_fullname}
    \bibliography{main}
}


\end{document}